\documentclass[letterpaper, 10 pt, conference]{ieeeconf}  

\IEEEoverridecommandlockouts                              

\usepackage{graphics} 
\usepackage{graphicx}
\usepackage{comment}
\usepackage{amsmath}
\usepackage{amssymb}
\usepackage[ruled,linesnumbered]{algorithm2e}
\usepackage{xcolor}

\RestyleAlgo{ruled}

\usepackage{verbatim} 
\usepackage[backend=biber,      
    style=ieee,
    bibstyle=ieee,              
    mincitenames=1,
    maxcitenames=2,
    sortcites=true,   
    giveninits=true,            
    backref=false
]{biblatex}
\title{\LARGE \bf
Robust Modeling and Controls for Racing on the Edge
}

\author{Joshua Spisak$^{1\text{*}}$, Andrew Saba$^{1\text{*}}$, Nayana Suvarna$^2$, Brian Mao$^{3}$, Chuan Tian Zhang$^{3}$, Chris Chang$^{4}$, \\ 
    Sebastian Scherer$^{1}$, and Deva Ramanan$^{1,5}$
\thanks{$^{1}$With Carnegie Mellon University
        {\tt\small jspisak, asaba, basti, deva @andrew.cmu.edu}}%
\thanks{$^{2}$With University of Pittsburgh
        {\tt\small nls71 @pitt.edu}}%
\thanks{$^{3}$With University of Waterloo
        {\tt\small bmao, ben.zhang @uwaterloo.ca}}%
\thanks{$^{4}$With Massachusetts Institute of Technology
        {\tt\small cwkchang @mit.edu}}%
\thanks{$^{5}$With Argo AI}%
\thanks{$\text{*}$Equal contribution}%
}

\begin{document}

\maketitle
\thispagestyle{empty}
\pagestyle{empty}

\begin{abstract}
    Race cars are routinely driven to the edge of their handling limits in dynamic scenarios well above 200mph. Similar challenges are posed in autonomous racing, where a software stack, instead of a human driver, interacts within a multi-agent environment. For an Autonomous Racing Vehicle (ARV), operating at the edge of handling limits and acting safely in these dynamic environments is still an unsolved problem. In this paper, we present a baseline controls stack for an ARV capable of operating safely up to 140mph. Additionally, limitations in the current approach are discussed to highlight the need for improved dynamics modeling and learning.

\end{abstract}

\section{INTRODUCTION}
    While some Autonomous Vehicles (AVs) have been deployed for public use, there are still edge cases that prevent safe usage in everyday life. It can take thousands or millions of hours of operations for these edge cases to be encountered. Autonomous racing is an opportunity to face several of these edge cases head-on. In particular, the Indy Autonomous Challenge (IAC) has held autonomous racing competitions such as the fastest single agent lap as well as a 1v1 passing competition. The IAC involves nine teams, making it the largest full-scale vehicle autonomous racing competition in the world. The competition is predicated on each team having the same hardware available, focusing the competition on algorithmic development to face the challenges of operating on the edge of handling limits and pushing autonomous capability to new levels. This paper outlines a controls software stack capable of racing up to 140mph. Later, we outline additional challenges and future opportunities to push high speed performance to the handling limits of the vehicle.
    
    \begin{figure}[t!]
    \centering
    \includegraphics[width=.9\linewidth]{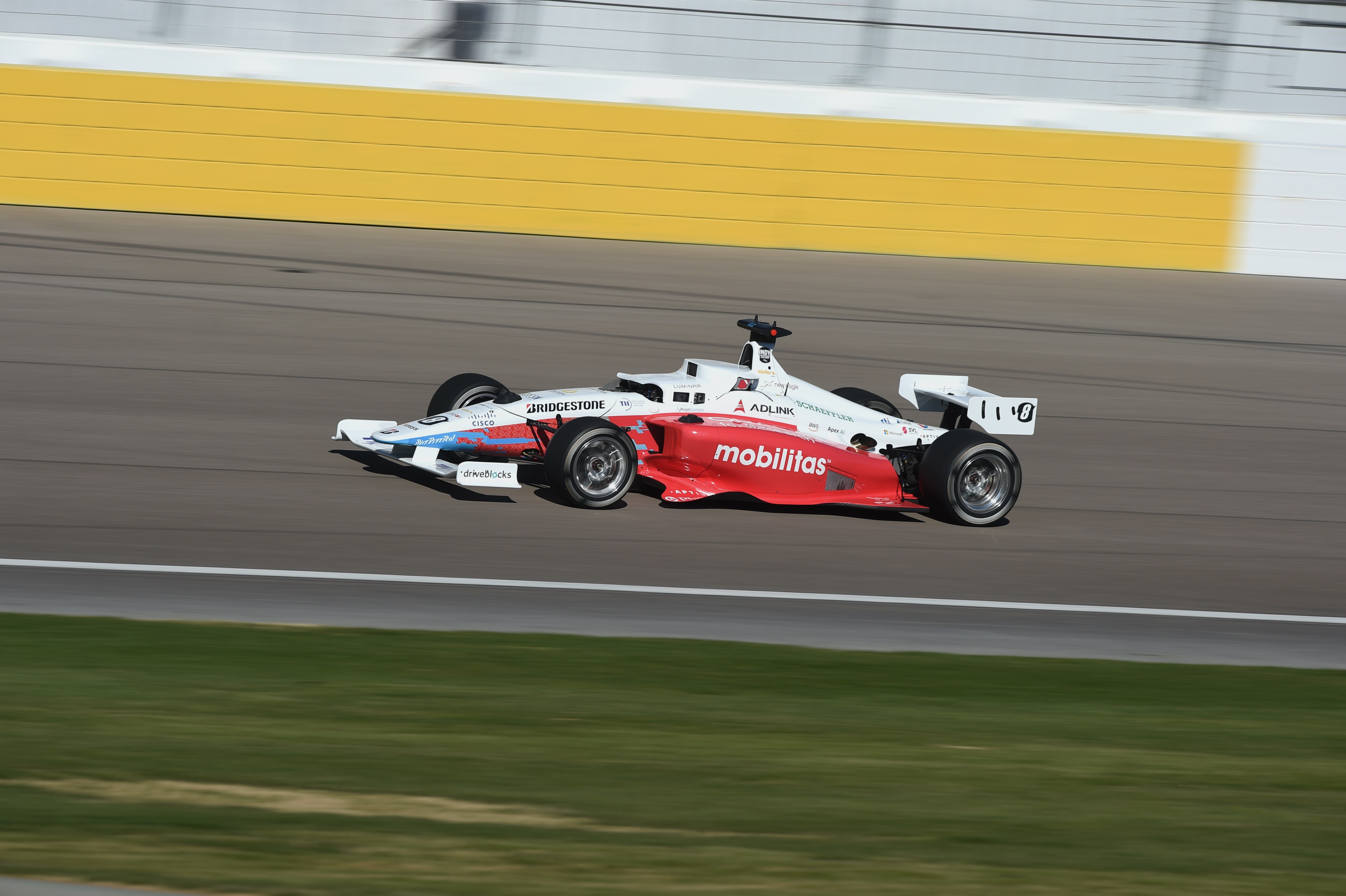}
       \caption{AV-21 driving at 137mph on the Las Vegas Motor Speedway}
    \label{fig:zoom}
    \end{figure}

\subsection{Problem}
    In autonomous racing, operating at high speeds is a minimum barrier to entry. Not only must the vehicle be able to move quickly, it must be able to react dynamically and stably. This can be a challenge even for human drivers, with multiple crashes occurring during the 2021 Indy 500 after drivers lost control of their vehicle. During a race, a human driver must be able to understand changing conditions such as track temperature, small pieces of tire rubber collecting on parts of the track, and gusts of wind disturbing the vehicle. They must also understand their own vehicle's changing conditions such as tire pressure, temperature, and wear. Changes in these conditions can cause loss of control and crashes. In addition, a safe ARV must maintain a reaction time that exceeds human capability --- in the time it takes a human driver to react, the vehicle can move several meters. These challenges express themselves in similar ways to autonomy software, where algorithms must be able to predict or react stably to these dynamic conditions quickly.

\subsection{Application to Urban Driving}
    Previous development in the autonomous vehicle space has been focused on the urban environment in which vehicles are at low dynamic loading. While racing vehicles move at speeds in excess of 200mph, the maximum speed limit in the United States is 85mph, and the maximum speed limit world-wide is 130mph. Most traffic moves slower than a racing vehicle; still, traffic related accidents are the number eight cause of deaths worldwide and a third of these deaths are caused by speeding \cite{road_safety}. Primary reasons for increased crash risk at higher speeds include a longer stopping distance, shorter reaction time, and bringing the vehicle to higher dynamic load than they might experience normally or be designed for making it more difficult to control.
    
    While urban vehicles do not experience the same dynamic load as racing vehicles do at 200mph+, they can approach similar limits, especially in adverse weather. They are also subject to non-negligible disturbances, such as aerodynamic effects at high speeds, potholes, and ice patches. Methods for stable high-speed control should be able to handle these same classes of disturbances and complex dynamics. Methods for autonomous racing can also push the limits of autonomous capability to minimize reaction time to dynamic adversarial agents (such as a drunk drivers) as well as minimize stopping distances to produce safer methods of transit with increased speed and efficiency.

\subsection{Content Overview}
    The results obtained in this paper were obtained using a fairly simple architecture. Our localization utilized the Novatel GNSS-INS system paired with robot localization \cite{robot_localization}. This, as well as a racing line, feed into a tracking controller. This paper briefly discusses an offline racing line generation method, but we will primarily focus on the tracking controller and an online modeling scheme that can feed parameters to this controller in an online manner.
    
    While the components used in this paper are not particularly novel, this specific formulation of the lateral controller combining pure-pursuit and LQR is, to the best of the authors' knowledge, unique and different from any prior formulations. The authors also believe there is value in demonstrating capability at high speeds with simple tracking frameworks, and describing the issues experienced during deployment as well as possible solutions.


\section{Racing Line Generation}\label{racingline}
    The racing line is represented spatially as a series of 2D points $(x, y)$. Due to the low road curvature of the oval speedways we race on, it suffices to decouple target speed from the racing line, since we assume that the same target speed is reasonable everywhere on the racing line as long as the target speed is reasonably slower than the true operating limits (over 200mph).
    
    We constrain the racing line to be a closed Spiro spline \cite{Levien09} interpolating a series of manually selected waypoints $(x, y)$. While the tracking controller is compatible with any reasonably smooth racing line, the Spiro spline is a particularly attractive spline family because:
    \begin{itemize}
        \item it exhibits $G^4$-continuity, meaning that the second derivative of curvature is continuous.
        \item it is an efficient approximation of the Minimum Variation Curve (MVC), which minimizes the integral of curvature rate, i.e. steering effort.
    \end{itemize}
    These properties make the Spiro spline particularly effective at minimizing instability in the downstream tracking controller.
    
    The manual selection of waypoints is up to user discretion and depends on the use case. For example, in order to generate a natural racing line (one where curvature is generally minimized), a good choice of waypoints would be the points where the vehicle is expected to drive closest to the boundaries of the track, i.e. the apexes. However, in order to generate a racing line that closely follows the center of the track, one might choose waypoints solely from the centerline. In order to ensure that the final racing line maintains a desired safety distance from the track boundaries, it may be necessary to tweak these waypoints and generate a new line multiple times.


\section{Tracking Controller}\label{approach}
\subsection{Overview}\label{tracking_overview}
\begin{figure}[t!]
    \centering
    \includegraphics[width=.9\linewidth]{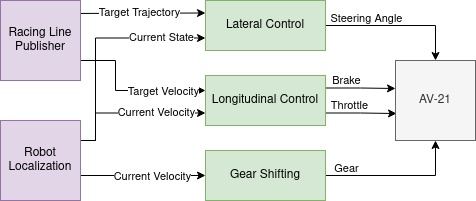}
       \caption{Stack and Controller Overview}
    \label{fig:stack}
    \end{figure}

    The tracking controller decomposes the tracking problem into three different problems: lateral tracking, longitudinal tracking, and gear shifting. Lateral tracking is concerned with generating a steering angle that will converge the vehicle to the path. This utilizes a combined pure-pursuit LQR control scheme where the LQR controller is evaluated relative to some look-ahead point. The longitudinal controller is concerned with maintaining a particular velocity. The gear shifting controller is concerned with selecting the best gear to maintain the current speed. Most of the content in this section will focus on the lateral tracking as that is the most interesting and novel portion of the stack. The stack's architecture is described in Figure \ref{fig:stack}.
    
\subsection{LQR}\label{sec:lqr}
    Given a continuous-time linear system of the form in Equation \ref{continuous_linear} we can define a quadratic cost function such as Equation \ref{cost_function} which is constrained by Equation \ref{continuous_linear}. Equation \ref{cost_function} also has the constraints that $Q = Q^T \geq 0$ and $R = R^T \geq 0$.

    We can derive an optimal feedback policy that minimizes the cost function in Equation \ref{cost_function} over the infinite horizon in time. Solving this problem is trivial and can done using Matlab's lqr function as well as many other available solvers such as ones detailed in \cite{lqr_solver_comparison}. This resulting feedback policy takes the form of Equation \ref{opt_feedback}.
    \begin{equation}\label{continuous_linear}
        \dot x(t) = A x(t) + B u(t)
    \end{equation}
    \begin{equation}\label{cost_function}\begin{split}
        J=\int_{t=0}^{\infty} \left[ x(t)^T Q x(t) + u(t)^T R u(t) \right]dt
    \end{split}\end{equation}
    \begin{equation}\label{opt_feedback}
        u^*(t) = K(t)x(t)
    \end{equation}

\subsection{Dynamic Bicycle Model}\label{sec:dbm}
    We utilize a four state dynamic bicycle model from \cite{Rajamani_2005}, shown in Equation \ref{eq:dbm}. As noted in \ref{tracking_overview} we decompose the problem into separate lateral and longitudinal controllers, the lateral controller reasons about the lateral position of the vehicle and the yaw of the vehicle $(y, \psi)$. The input for this system is steering angle $\delta$, and it assumes some constant velocity $V_x$. Other parameters of the system include $C_{\alpha f}$, $C_{\alpha r}$ (cornering stiffness of the front and rear tires respectively), $l_f$, $l_r$ (length of the vehicle from the center for gravity to the front and rear axles respectively). The parameter $I_z$ is the moment of inertia about the z axis, and $m$ is the mass of the vehicle.

    \begin{equation}\label{eq:dbm}\begin{split}
        & A = \\
        & \begin{bmatrix}
            0 & 1 & 0 & 0 \\
            0 & -\frac{2 C_{\alpha f} + 2 C_{\alpha r}}{m V_x} & 0 & -V_x - \frac{2 C_{\alpha f} l_f - 2 C_{\alpha r} l_r}{m V_x} \\
            0 & 0 & 0 & 1 \\
            0 & -\frac{2 l_f C_{\alpha f} - 2 l_r C_{\alpha r}}{I_z V_x} & 0 & -\frac{2 l_f^2 C_{\alpha f} + 2 l_r^2 C_{\alpha r}}{I_z V_x}
        \end{bmatrix} \\
        & \begin{bmatrix}
            \dot y \\ \ddot y \\ \dot \psi \\ \ddot \psi 
        \end{bmatrix} = A \begin{bmatrix}
            y \\ \dot y \\ \psi \\ \dot \psi 
        \end{bmatrix} + \begin{bmatrix}
            0 \\
            \frac{2 C_{\alpha f}}{m} \\
            0 \\
            \frac{2 l_f C_{\alpha f}}{I_z}
        \end{bmatrix} \delta
    \end{split}\end{equation}

    To utilize this model with the LQR formulation above, the model is rephrased such that the state space is error with respect to a target trajectory as shown in Equation \ref{rep_dbm}. The error term $e_1$ is the lateral error of the vehicle from the target trajectory and $e_2$ being the yaw error of the vehicle relative to the target trajectory.

    \begin{equation}\label{rep_dbm}\begin{split}
        & A = \\ & \begin{bmatrix}
            0 & 1 & 0 & 0 \\
            0 & -\frac{2 C_{\alpha f} + 2 C_{\alpha r}}{m V_x} & \frac{2 C_{\alpha f} + 2 C_{\alpha r}}{m} & - \frac{2 C_{\alpha f} l_f - 2 C_{\alpha r} l_r}{m V_x} \\
            0 & 0 & 0 & 1 \\
            0 & -\frac{2 l_f C_{\alpha f} - 2 l_r C_{\alpha r}}{I_z V_x} & \frac{2 l_f C_{\alpha f} - 2 l_r C_{\alpha r}}{I_z} & -\frac{2 l_f^2 C_{\alpha f} + 2 l_r^2 C_{\alpha r}}{I_z V_x}
        \end{bmatrix} \\
        & \begin{bmatrix}
            \dot e_1 \\ \ddot e_1 \\ \dot e_2 \\ \ddot e_2
        \end{bmatrix} = A \begin{bmatrix}
            e_1 \\ \dot e_1 \\ e_2 \\ \dot e_2
        \end{bmatrix} + \begin{bmatrix}
            0 \\
            \frac{2 C_{\alpha f}}{m} \\
            0 \\
            \frac{2 l_f C_{\alpha f}}{I_z}
        \end{bmatrix} \delta
    \end{split}\end{equation}

    Given a target at some point along the trajectory with a given $\begin{bmatrix} x^* & y^* & \psi^* & \dot \psi^* \end{bmatrix}$ as well as the vehicle position $\begin{bmatrix} x & y & \dot x & \dot y & \psi & \dot \psi \end{bmatrix}$, with all states specified in some inertial frame excluding $\dot x$ and $\dot y$ which is the velocity of the vehicle in the body frame, we can calculate the error state as follows:
    \begin{equation}\label{error_terms}
        \begin{bmatrix}
            e_1 \\
            \dot e_1 \\
            e_2 \\
            \dot e_2
        \end{bmatrix} = \begin{bmatrix}
            (x^* - x) \sin({-\psi^*}) + (y^* - y) \cos({-\psi^*}) \\
            \dot y + \dot x (\psi - \psi^*) \\
            \psi - \psi^* \\
            \dot \psi - \dot \psi^*
        \end{bmatrix}
    \end{equation}
    
    This model makes several assumptions, including small angle assumptions on the steering angle which hold well on ellipsoidal tracks where we command a maximum steering angle of approximately $0.1$ [rad]. The model also assumes constant velocity. A reformulation to a linear time-varying model could allow for varying velocity, however for simplicity we accept that assumption then mitigate this with a series of feedback controllers derived at different velocities, this is described in Section \ref{sec:pp_lqr}. For greater detail on this formulation of the dynamics see \cite{Rajamani_2005}.

\subsection{Pure Pursuit, LQR Tracking Algorithm}\label{sec:pp_lqr}
    Given the model formulation in \ref{sec:dbm} and the LQR formulation in \ref{sec:lqr} we can make a lateral tracking algorithm by combining a pure-pursuit style look-ahead point and a feedback mechanism like in Equation \ref{opt_feedback} generated using LQR. The look-ahead point is queried simply by looking for the point on the trajectory that is ahead of the vehicle a given distance $d$ away. To generate the feedback policy lqr requires an $A$, $B$, $Q$ and $R$ matrix. The $A$ and $B$ matrix can be obtained using the dynamics from Equation \ref{rep_dbm}. The $Q$ and $R$ matrices can be obtained empirically where the trace of the $Q$ matrix defines the weights on the error terms in Equation \ref{error_terms}, the $R$ matrix defines the weight on the steering angle essentially allowing us to dampen the steering.
    
    To account for varying dynamics at different speeds we have the concept of a velocity bracket ranging $[\dot x_{\text{low}}, \dot x_{\text{high}})$ such that if the vehicles current velocity is within that bracket it utilizes a particular parameter set $p = (Q_{(0,0)}, Q_{(1,1)}, Q_{(2,2)}, Q_{(3,3)}, R_{(0,0)}, R_{(1,1)})$. The algorithm inputs a series of brackets which are assumed to cover the range of velocities we expect to cover or generally the range $[0, \infty)$ with no overlap between them. The look-ahead distance is also varied a function of speed where $d = d_{\text{base}} + k_{v,d} \dot x$. These parameter sets can be defined empirically; however, we mostly tuned the dampening on the steering angle, increasing it with speed.
    
    The algorithm runs as follows. First, it initializes by loading in all velocity brackets and the associated parameters. It then generates the feedback policies $K$ for each bracket using the average velocity of the bracket (unless one of the bounds is $\infty$ in which case it uses the lower velocity). Then at each time step, given the current state of the vehicle and a target trajectory it performs a look-ahead query on the target trajectory to get the goal point. Using this goal point and the current state an error state can be generated. Finally, with the current vehicle velocity, it queries the velocity brackets for the relevant feedback policy which is applied on the error state to get the optimal steering angle. This algorithm is summarized in Algorithm \ref{alg:tracking}.
    
    \SetKwComment{Comment}{/* }{ */}
    \begin{algorithm}
    \SetAlgoLined
    \caption{Lateral Tracking Algorithm}\label{alg:tracking}
    $P \gets \{(v_{1, \text{low}}, v_{2, \text{low}}, K_1), \hdots, (v_{n, \text{low}}, \infty, K_n)\}$ \Comment*[r]{We assume parameters have been loaded and the feedback policies generated.}
    $d_{\text{base}}, k_{v,d} \gets \text{loadParams}()$\;
    \While{$true$}{
        $x, y, \dot x, \dot y, \psi, \dot \psi \gets \text{getState}()$\;
        $\tau \gets \text{getTrajectory}()$\;
        $d \gets d_{\text{base}} + k_{v,d} * \dot x$\;
        $K \gets \text{getFeedbackMatrix}(\dot x, P)$\;
        $x^*, y^*, \psi^*, \dot \psi^* \gets \text{lookahead}(x, y, d, \tau)$\;
        $e \gets \text{errorMatrix}$\ $(x, y, \dot x, \dot y, \psi, \dot \psi, x^*, y^*, \psi^*, \dot \psi^*)$\;
        $u \gets -K e$\;
        $\text{commandSteering}(u)$\;
    }
    \end{algorithm}

\subsection{Longitudinal Control and Gear Shifting}
    The throttle and brake control utilizes a simple P control scheme paired with a feed-forward term to account for drag. This is expressed with in Algorithm \ref{alg:long} line \ref{alg:pid_line}, which returns a command value. This command is then interpreted as either a throttle or brake command depending on whether the command is positive or negative. Supposing the command is negative we also apply a scaling factor $\alpha_{\text{brake}}$ to the brake, tuned assuming there is no scaling factor from the command to throttle. We also apply smoothing on the throttle and brake to prevent instantaneous acceleration or deceleration that could make the vehicle unstable, parameterized simply by an allowed rate of change with respect to time. This pipeline is summarized in Algorithm \ref{alg:long}.

    The gear shifting strategy is based on a simple look up table where given the current velocity of the vehicle we can compute the optimal gearing. This gearing table was generated based on a model of the engine and track parameters to maximize torque, the computation of which is outside the scope of this paper.

    \begin{algorithm}
    \caption{Longitudinal Tracking Algorithm}\label{alg:long}
    $k_p, k_{\text{feed forward}}, \alpha_{\text{brake}}, \delta_{\text{throttle}}, \delta_{\text{brake}} \gets \text{loadParams}()$\;
    $\text{throttle\_previous}, \text{brake\_previous} \gets 0$\;
    \While{$true$}{
        $x, y, \dot x, \dot y, \psi, \dot \psi \gets \text{getState}()$\;
        $v_{\text{target}} \gets \text{getTarget}()$\;
        $\text{command} \gets k_p (v_{\text{target}} - \dot x) + k_{\text{feed forward}} * v_{\text{target}}$\; \label{alg:pid_line}
        $\text{throttle}, \text{brake} \gets 0$
        \eIf{$\text{command} \geq 0$}{
            $\text{throttle} \gets \text{command}$\;
        }{
            $\text{brake} \gets - \alpha_{\text{brake}}\text{command}$\;
        }
        $\text{throttle} \gets \text{smooth}(\text{throttle\_previous}, \text{throttle}, \delta_{\text{throttle}})$\;
        $\text{brake} \gets \text{smooth}(\text{brake\_previous}, \text{brake}, \delta_{\text{brake}})$\;
        $\text{comamnd}(\text{throttle}, \text{brake})$\;
        $\text{throttle\_previous} \gets \text{throttle}$\;
        $\text{brake\_previous} \gets \text{brake}$\;
    }
    \end{algorithm}

\subsection{Results}
    
\begin{figure}[t!]
    \centering
    \includegraphics[width=.9\linewidth]{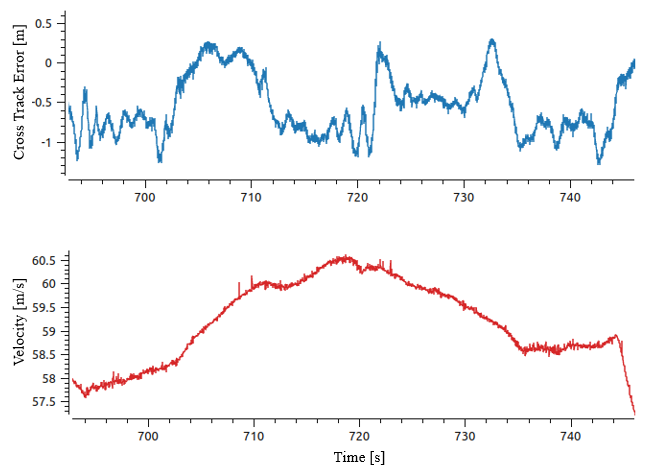}
      \caption{Top plot: cross-track error (CTE). Lower plot: Vehicle Velocity}
    \label{fig:cte_plots}
    \end{figure}

\begin{figure}[t!]
    \centering
    \includegraphics[width=.9\linewidth]{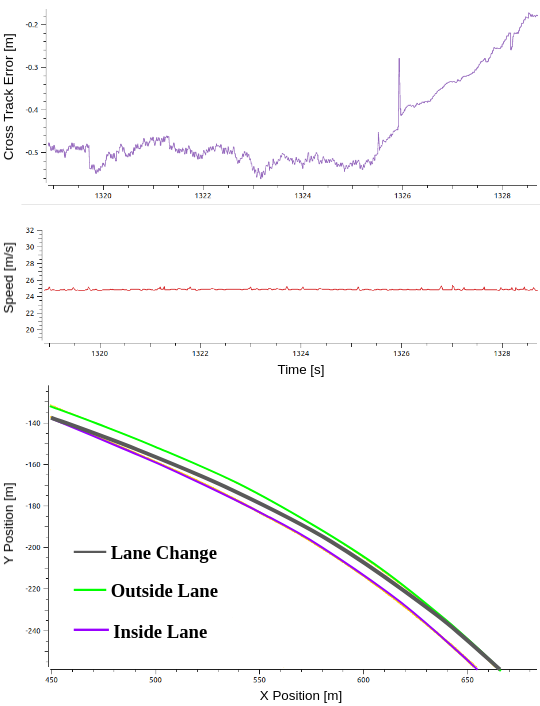}
      \caption{Top plot: cross-track error (CTE). Middle plot: Vehicle Velocity, Lower plot: lanes and lane change maneuver.}
    \label{fig:lane_change_plots}
    \end{figure}

    The stack shown in Figure \ref{fig:stack} was evaluated over several performance laps at the Las Vegas Motor Speedway with velocities ranging between 25 [m/s] and 60.5 [m/s]. We experienced the worst performance at the highest speed targeting 60 [m/s]. We plot velocity and cross-track error (CTE) for this portion of the evaluation in Figure \ref{fig:cte_plots}. The maximum CTE experience was 1.3 [m], experienced around bends, while the mean absolute CTE was 0.42 [m]. We also evaluate the controller on a lane-change task on the same track at 25 [m/s]. The results are shown in Figure \ref{fig:lane_change_plots}. The maximum CTE experienced was 0.55 [m].
    
\subsection{Discussion}
    
    
    In this section we present a simple method to combine pure-pursuit and an optimal-control method utilizing LQR. While not a particularly complex optimal-control method, it is able to stabilize the vehicle at high speeds with reasonable cross-track error. We consider this simplicity to be a strength of this algorithm. Classical controllers such as those described in \cite{old_tracking_methods} are still utilized in autonomous racing as seen in \cite{hartmann2021autonomous} which utilizes a pure-pursuit method as the low-level controller and \cite{sukhil2021adaptive} which evaluates an adaptive pure-pursuit method. However, more commonly Model Predictive Control methods have been utilized which can have the benefits of reasoning more explicitly about effects such as drag and velocity control over time as in \cite{jung2021game}, or to further reason about dynamic limits and uncertainty as in \cite{wischnewski2022tube}. 
    
    Our approach meets these somewhere in the middle: we utilize a simple optimization scheme that yields a feedback controller that can reason about vehicle dynamics without employing receding horizon optimization. This results in strongly guaranteed fast execution time, and stability at high speeds as demonstrated in the results on vehicle at 140 [mph]. However, there are some shortcomings in this approach in the empirical tuning of the look-ahead parameters which can have a strong effect on performance. Future work could include incorporating an optimized look-ahead distance such as in \cite{lookahead_adjust1} or \cite{sukhil2021adaptive}. Additionally this method does not truly incorporate the ability to handle changing conditions outside of feedback once disturbances occur. In Section \ref{opp_improve} we discuss further opportunities to improve this and other control frameworks.

\section{Opportunities for Improvement}\label{opp_improve}
    Non-linear dynamics at the edge of vehicle handling necessitate complicated control regimes that still fall short of matching and exceeding human driver performance \cite{human_comparison}. While in \cite{Srinivasan_2021} the authors claim performance that exceeds a human driver, it must be noted that the ARV was driving without the additional weight of a human and was limited to speeds of up to 18m/s, whereas larger ARVs are racing at speeds in excess of 77 [m/s] (173 [mph]) \cite{iac_article}.
    
    In the following sections, we will discuss the vehicle dynamics problems in more detail and present a modeling and dynamics learning pipeline that can address these challenges.
    
    \begin{figure}[t!]
    \centering
    \includegraphics[width=.9\linewidth]{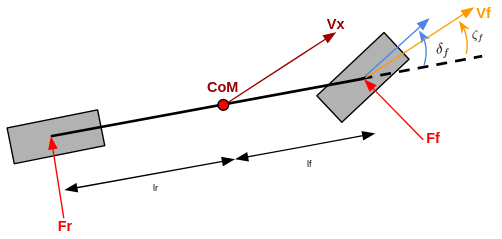}
      \caption{Dynamic Bicycle Model. $l_f$ and $l_r$ denote the distance between the front and rear tires and the center of mass (CoM). $V_x$ and $V_f$ denote the velocity at the CoM and front tire, respectively. $\zeta_f$ is the direction of the front wheel velocity and $\delta_f$ is the current steering angle. $F_r$ and $F_f$ are the lateral forces acting on the rear and front tires, respectively.}
    \label{fig:dbm}
    \end{figure}

\subsection{Online Tire Modeling and Learning}
    At the core of our controller is the dynamic bicycle model, a portion of which is shown in Figure \ref{fig:dbm}. This model, for the center of mass and for the front and rear tires, is centered around three critical angles: $\zeta$, or the direction of the velocity vector; $\delta$, or the current angle the tire or vehicle body is pointing; and $\alpha$, or slip angle, which is defined as $\alpha = \delta - \zeta$.
    
    The slip angle is what is ultimately used to estimate the lateral forces acting on the tire. The Pacejka tire model is one of the most widely used empirical tire models due to its accuracy when compared to empirically measured data. Tire forces are expressed as:
    
    \begin{equation}
        F = D \sin \bigg( C \arctan \big(B\alpha - E(B \alpha - \arctan(B \alpha) \big)\bigg) \mu F_z
        \label{Pacejka_tire_model}
    \end{equation}
    
    where $B$ is the stiffness factor, $C$ is the shape factor, $D$ is the peak value, $E$ is the curvature factor, $\alpha$ is the slip angle, $\mu$ is the tire road friction coefficient, and $F_z$ is the vertical tire force. 
    
    For computational costs and simplicity's sake, our controller utilizes a linear approximation of the Pacejka tire model, which defines a cornering stiffness $C_r$ and $C_f$ for the rear and front tires, respectively. With this, the lateral forces are now defined as $F_r = C_r * \alpha_r \label{equ:linear_tire_modelR}$ and $F_f = C_f * \alpha_f \label{equ:linear_tire_modelF}$.
    
    With our full state estimate and estimated slip angle, we are capable of estimating the linear model parameters, are not static and change over time due to tire degradation, temperature, pressure, and other factors. This degradation is even more severe under the context of racing where tires are often pushed to their friction limits. Therefore, it would be beneficial to develop an estimation methodology to determine tire model parameters online.
    
    \begin{figure}[t!]
    \centering
    \includegraphics[width=.9\linewidth]{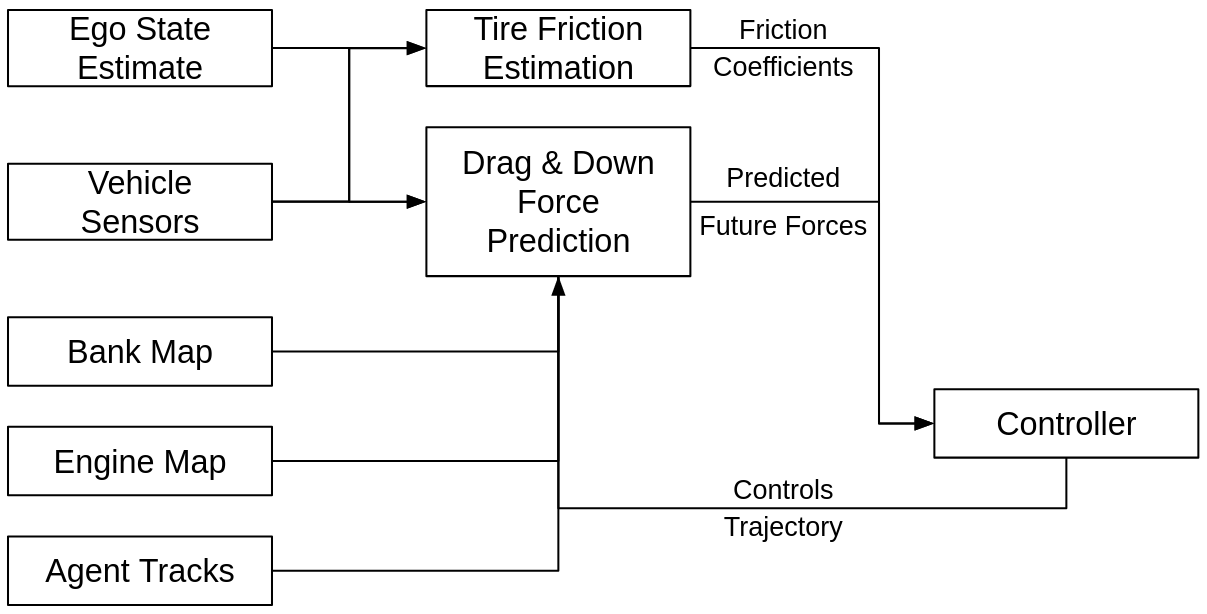}
      \caption{Proposed modeling pipeline. Vehicle state and sensor input is used to estimate the friction coefficients at the current time step. A separate module uses vehicle state and sensors, track banking, and other agents locations to adapt a model that predicts the drag and down forces on the vehicle. Since drag and down forces can be estimated, this model can be trained in a self-supervised fashion. Finally, this model is used by the controller to optimize the candidate trajectories and predict future vehicle dynamics.}
    \label{fig:modeling_pipeline}
    \end{figure}
    
    Similar ideas have been introduced before, including in \cite{tum-friction} and others. However, tire friction is only one part in a larger dynamics model. For example, estimating friction does not capture how wind or aerodynamic effects from other vehicles impact performance. The authors in \cite{gpr_amz} address this with a more holistic approach, using a Gaussian Process Regression (GPR) model that is trained online to estimate the error in the vehicle model. This allows for their controller to reason about the overall dynamic state of the vehicle, capturing more than just changes in friction potential from the tires.
    

\subsection{Aerodynamic Force Learning and Prediction}
    In addition to two RTK-GNSS systems, the AV-21 platform is equipped with a wide range of sensors that capture information about the tires, suspension, engine, and vehicle loading. Instead of predicting total model error, a better approach may be to predict individual components of the dynamics model (see Figure \ref{fig:modeling_pipeline}). At the core of the proposed approach is a module that learns to predict drag and down forces, given a trajectory of controls. The model is trained online asynchronously, with the vehicle state, sensors, bank map of the track, engine torque curve, and current tracks of other agents as inputs. Since drag and down force can be estimated online from sensor data and the vehicle state is continuously measured with high accuracy RTK-GPS, the model can be trained online in a self supervised fashion. As the controller optimizes the target trajectory, it can utilize use the learned model's predictions.
    
    Through online adaptation, this approach is robust to novel scenarios and conditions. Aerodynamics are notoriously difficult to model and predict accurately, so online learning is necessary to adapt to the environment and the current state of the world. With this adaptation and learning, the controller can become more aggressive, pushing the envelope further.


\section{Conclusion}
    We have presented a baseline controls stack for an autonomous race vehicle (ARV). This stack, while simple, provides a robust baseline to expand and build upon moving forward in future competitions and technical developments. Additionally, we have also presented a concept modeling approach that can utilize vehicle and environment information to better predict how vehicle dynamics will unroll. Finally, future work will be focused on implementing and testing this approach, to prepare for future races.

\section*{ACKNOWLEDGMENT}
    Andrew Saba and Deva Ramanan are supported by the CMU Argo AI Center for Autonomous Vehicle Research.
    
    Thank you to Matthew Travers for his continually helpful advice and support. Finally, thank you to the entire MIT-Pitt-RW Indy Autonomous Challenge team and team sponsors for supporting the deployment of this work.

\printbibliography

\end{document}